\documentclass[pmlr]{jmlr} 

\newcommand{\tagdsmode}{proceedings}

\makeatletter
\newcommand{\tagdssubmission}{submission}
\newcommand{\tagdsproceedings}{proceedings}

\ifx\tagdsmode\tagdsproceedings

\else\ifx\tagdsmode\tagdssubmission
  \def\ps@jmlrtps{%
    \let\@mkboth\@gobbletwo
    \def\@oddhead{\scriptsize Under Review at the 2nd Conference on Topology, Algebra, and Geometry in Data Science\hfill}%
    \let\@evenhead\@oddhead
    \def\@oddfoot{}%
    \let\@evenfoot\@oddfoot
  }

\else
  \def\ps@jmlrtps{%
    \let\@mkboth\@gobbletwo
    \def\@oddhead{}%
    \let\@evenhead\@oddhead
    \def\@oddfoot{}%
    \let\@evenfoot\@oddfoot
  }
\fi\fi
\makeatother

\usepackage{longtable}

\usepackage{booktabs}
\usepackage[load-configurations=version-1]{siunitx} 

\theorembodyfont{\upshape}
\theoremheaderfont{\scshape}
\theorempostheader{:}
\theoremsep{\newline}

\jmlrvolume{334}
\jmlryear{2026}
\jmlrworkshop{Topology, Algebra, and Geometry in Data Science}

\title[Look Before You Lift: TopoExplorer]{Look Before You Lift: Visual and Quantitative Diagnostics for Topological Deep Learning}

\ifx\tagdsmode\tagdssubmission

\else

\author{
\Name{Mathilde Papillon$^*$}
\Email{papillon@ucsb.edu}\\
\Name{Guillermo {Bern\'ardez$^*$}}
\Email{guillermo\_bernardez@ucsb.edu}\\
\Name{{\'A}lvaro {Ball\'on Barreiro}}\\
\addr University of California Santa Barbara \\
\AND
\Name{Marco Montagna}\\
\addr{Sapienza Universit\'a di Roma} \\
\AND
\Name{R\'emi Devaux}\\
\Name{Antoine Jardin}\\
\addr Arlequin AI
\AND
\Name{Nina Miolane}\\
\addr University of California Santa Barbara
}

\fi

\ifx\tagdsmode\tagdsproceedings
\fi

\usepackage{amsmath,amsfonts,bm}

\def\eqref#1{equation~\ref{#1}}

\def\1{\bm{1}}

\DeclareMathAlphabet{\mathsfit}{\encodingdefault}{\sfdefault}{m}{sl}
\SetMathAlphabet{\mathsfit}{bold}{\encodingdefault}{\sfdefault}{bx}{n}

\def\gA{{\mathcal{A}}}

\def\gC{{\mathcal{C}}}

\def\gI{{\mathcal{I}}}

\def\gN{{\mathcal{N}}}

\def\gT{{\mathcal{T}}}

\usepackage[utf8]{inputenc}
\usepackage[T1]{fontenc}

\usepackage{url}
\usepackage{booktabs}
\usepackage{amsfonts}
\usepackage{nicefrac}
\usepackage{microtype}
\usepackage{xcolor}
\usepackage{amsmath}
\usepackage{amssymb}
\usepackage{graphicx}
\usepackage{tikz}
\usetikzlibrary{arrows.meta,positioning,fit,backgrounds}
\usepackage{tcolorbox}
\usepackage{enumitem}
\usepackage{float}
\usepackage{hyperref}
\usepackage{rotating}

\makeatletter
\newcommand\blfootnote[1]{%
  \begingroup
  \def\@makefntext##1{\noindent ##1}%
  \footnotetext{#1}%
  \endgroup
}
\makeatother

\begin{document}

\maketitle

\maketitle

\blfootnote{$^*$Equal contribution and co-corresponding authors.}

\begin{abstract}
Topological deep learning (TDL) methods rely on lifting raw data into higher-order discrete domains such as simplicial complexes, cell complexes, and hypergraphs. In practice, this lifting step is often treated as a black box: practitioners select a lifting and then tune architectures, with limited visibility into whether the induced higher-order connectivity is meaningful for the downstream task. To address this missing diagnostic layer, we propose a visualization technique called TopoExplorer that leverages the strictly augmented Hasse graph form of topological datasets for exploratory data analysis. For the first time, practitioners can easily visualize the incidence- and adjacency-based neighborhoods that define the lifted dataset, as well as read off key graph metrics that describe its structural and feature landscape. Via an extensive set of experiments across many datasets and liftings, we show that several of these metrics correlate with downstream model performance, suggesting they can help inform TDL preprocessing design. Our perspective reframes the TDL workflow from \emph{lift-train} to \emph{lift-look-design-train}, enabling more principled, interpretable, and efficient model development. TopoExplorer is hosted  at \href{https://topoexplorer.pagekite.me/}{topoexplorer.pagekite.me/}, and its source code is available at \href{https://github.com/geometric-intelligence/topoexplorer}{github.com/geometric-intelligence/topoexplorer}.
\end{abstract}


\section{Introduction}
Many real-world systems involve interactions among groups that go beyond pairwise relationships, like atoms in a molecular ring or groups of friends in a social network. Topological Deep Learning (TDL)~\citep{bodnar2023topological,hajij2023topological} generalizes graph neural networks (GNNs) to model and learn from these higher-order interactions. In this work, we specifically refer to TDL in terms of methods that learn on discrete higher-order domains; methods that use other tools to consider data topology, such as persistent homology, fall outside our scope.

To do so, many TDL pipelines begin with conventional data representations, such as point clouds or graphs, and apply a \emph{lifting procedure}. Intuitively, lifting acts as a structural scaffolding step: it uses connectivity rules or feature-based criteria to explicitly construct higher-order elements, such as mapping cliques of edges to filled simplicial faces. A TDL model is then trained on the resulting topological domain. While this workflow has produced promising empirical results, it also introduces a fundamental source of uncertainty: \emph{what structure did the lifting actually create, and does it reflect the relational patterns that matter for the task?}

In many practical settings, lifting choices are made from a menu of available constructions, often with minimal structural inspection. This issue grows as toolkits make more lifting families available across simplicial, cellular, and hypergraph domains. For instance, TopoBench~\citep{telyatnikov2024topobench} has demonstrated the breadth of this design space through a systematic benchmark of liftings and tasks across domains. Yet breadth also increases ambiguity: different liftings can induce drastically different connectivity patterns, even on the same dataset.
We argue that TDL needs a first-class interface between lifting and learning. Strictly augmented Hasse graph decompositions (introduced in \citet{papillon2025topotune}) provide such an interface by making incidence- and adjacency-based neighborhoods explicit in graph form, enabling practitioners to inspect how connectivity propagates across cell ranks before training. This is particularly relevant for neighborhood-driven architectures such as TopoTune~\citep{papillon2025topotune} and HOPSE~\citep{bernardez2026hopse}. By characterizing lifted connectivity a priori, such an interface would enable practitioners in making better informed lifting and neighborhood design choices.

\paragraph{Contributions.} We introduce \textbf{TopoExplorer}, a visual and quantitative diagnostic framework grounded in strictly augmented Hasse graph decompositions. The interactive app is hosted at \href{https://topoexplorer.pagekite.me/}{topoexplorer.pagekite.me/} and made available open-source at \href{https://github.com/geometric-intelligence/topoexplorer}{github.com/geometric-intelligence/topoexplorer}. Beyond mere visualization, TopoExplorer allows practitioners to extract key structural and feature-based graph metrics from the lifted complex; it reframes the standard TDL workflow from a blind \emph{lift-train} approach to a more principled \emph{lift-look-design-train} approach. To the best of our knowledge, this is the first exploratory data and visualization framework for better informing lifting design decisions in TDL. Through the analysis of extensive experiments using the default liftings in TopoBench, we show that some pre-training metrics computed in TopoExplorer correlate with downstream model performance, demonstrating the usefulness of such a structure-aware design pipeline. 

\section{Background}

This section introduces key concepts underpinning TDL. For brevity and generality, the following discussion centers on combinatorial complexes, which subsume all of the discrete topological domains leveraged by TDL~\citep{hajij2023topological}. 

\paragraph{Combinatorial Complex.} A \emph{combinatorial complex} is a triple $(\mathcal{V}, \mathcal{C}, \textrm{rk})$ consisting of a set $\mathcal{V}$, a subset $\mathcal{C}$ of the powerset $\mathcal{P}(\mathcal{V}) \backslash\{\emptyset\}$, and a rank function $\textrm{rk}: \mathcal{C} \rightarrow \mathbb{Z}_{\geq 0}$ with the following properties:
\begin{enumerate}[leftmargin=20pt,itemsep=1pt,parsep=0pt,topsep=0pt]
    \item for all $v \in \mathcal{V},\{v\} \in \mathcal{C}$ and $\textrm{rk}(\{v\})=0$;
    \item the function $\textrm{rk}$ is order-preserving, i.e., if $\sigma, \tau \in \mathcal{C}$ satisfy $\sigma \subseteq \tau$, then $\textrm{rk}(\sigma) \leq$ $\textrm{rk}(\tau)$.
\end{enumerate}
The elements of $\mathcal{V}$ are the nodes, while the elements of $\mathcal{C}$ are called cells (i.e., group of nodes). 
The rank of a cell $\sigma \in \mathcal{C}$ is $k:=\textrm{rk}(\sigma)$, and we call it a $k$-cell. $\mathcal{C}$ simplifies notation for $(\mathcal{V}, \mathcal{C}, \textrm{rk})$, and its dimension is defined as the maximal rank among its cells: $\mathrm{dim}(\mathcal{C}):= \max_{\sigma \in \mathcal{C}} \textrm{rk}(\sigma)$. 

\begin{figure}[t]
    \vspace{-20pt}
    \centering
    \includegraphics[width=\textwidth]{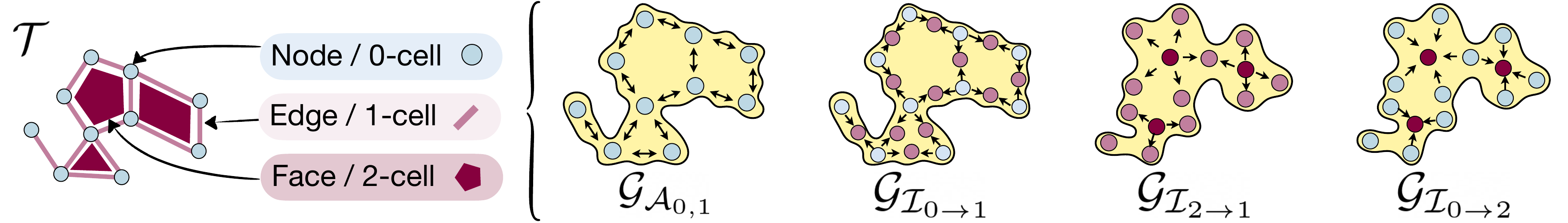}
\caption{\textbf{Neighborhood expansions of a complex.} Given a complex $\gT$ (left), three examples of strictly augmented Hasse graphs $\mathcal{G}_\mathcal{N}$  corresponding to 4 neighborhood functions: adjacency of nodes w.r.t. edges ($\gA_{0,1}$), incidence from nodes to edges ($\gI_{0\to 1}$), incidence from faces to nodes ($\gI_{2\to 1}$), and incidence from nodes to faces ($\gI_{0\to 2}$). Adopted from \citet{bernardez2026hopse}} 
\label{fig:background}
\vspace{-10pt}
\end{figure}

\paragraph{Neighborhood Relations: Adjacencies and Incidences.} Combinatorial complexes can be equipped with a notion of neighborhood among cells that induces  topological structures \footnote{Formally, denoting by $\mathcal{P}(\cdot)$ the power set, a neighborhood function $\mathcal{N}: \mathcal{C} \rightarrow \mathcal{P}(\mathcal{C})$ on a complex $\gT$ maps each cell $\sigma \in \mathcal{C}$ to a collection of \emph{neighbor cells} $\mathcal{N}(\sigma) \subset \mathcal{C}$.}. 
Traditional neighborhood functions include \emph{adjacencies} ($\mathcal{A}_{t,s}$), connecting cells of the same rank $t$ ---via relations w.r.t. cells of another rank $s$--- and \emph{incidences} ($\mathcal{I}_{s\to t}$), linking cells of different ranks from a source rank $s$ to a target rank $t$ (see Figure \ref{fig:background} for some visualizations, Appendix \ref{app:nbhds} for full definitions). 


\paragraph{Strictly augmented Hasse graphs.} Given a complex $\gT$, each particular neighborhood function $\mathcal{N}$ induces a strictly augmented Hasse graph $\mathcal{G}_{\mathcal{N}} = (\mathcal{C}_\mathcal{N}, \mathcal{E}_\mathcal{N})$ \citep{papillon2025topotune}, defined as the directed graph whose nodes and edges are given, respectively, by 
\begin{equation}\label{eq:hasse}
\mathcal{C}_{\gN} = \{ \sigma \in \gC \mid \gN(\sigma) \neq \emptyset \} 
\quad \text{and} \quad
\mathcal{E}_{\gN} = \{ (\tau, \sigma) \mid \sigma, \tau \in \gC_{\gN},\ \tau \in \gN(\sigma) \}.
\end{equation}
Figure \ref{fig:background} visually shows some examples of the graph expansions induced by these neighborhood functions. Notably, \emph{strictly} comes from the fact that the set of nodes of $\mathcal{G}_{\gN}$ is restricted to $\mathcal{C}_{\gN}$, i.e., to those cells that have at least one neighbor according to $\gN$. 

\paragraph{Lifting.} A lifting transformation is a map $\mathcal{L}: \mathcal{D}_{\text{in}} \to \mathcal{D}_{\text{out}}$ between two discrete topological domains, defined by a fixed rule applied uniformly to the input domain. While traditionally conceptualized as promoting lower-order data (e.g., a simple graph) to a higher-order complex, $\mathcal{L}$ more broadly defines any structural translation between topological spaces. Formally, given an input domain $\mathcal{D}_{\text{in}} = (V_{\text{in}}, \mathcal{C}_{\text{in}}, \text{rk}_{\text{in}})$, the lifting function identifies specific structural motifs or sub-patterns $\mathcal{P} \subseteq \mathcal{C}_{\text{in}}$ and maps them to cells in a target complex $\mathcal{D}_{\text{out}} = (V_{\text{out}}, \mathcal{C}_{\text{out}}, \text{rk}_{\text{out}})$. A ubiquitous example in graph-based TDL is the clique lifting, where $\mathcal{L}$ maps $(k+1)$-cliques in $\mathcal{D}_{\text{in}}$ to $k$-cells in $\mathcal{D}_{\text{out}}$. However, this flexible formulation equally supports transformations between domains of the same dimension or mappings from higher to lower-order representations (e.g., skeletal projections). 

\section{Related Work}

\paragraph{Higher-Order Representation Learning}
Topological Deep Learning (TDL), as surveyed in \citet{papillon2023architectures}, has emerged as a generalization of Graph Neural Networks (GNNs), extending learning from pairwise edges to higher-order interactions. Early work focused on Simplicial Neural Networks (SNNs) \citep{ebli2020simplicial} and Cell Complex Neural Networks (CCNNs) \citep{bodnar2021weisfeiler}, which leverage the rigid hierarchy of algebraic topology. All of these works were unified under the umbrella of Combinatorial Complexes (CCs) \citep{hajij2024topologicalbook}. Modern TDL architectures are increasingly modular, defined by neighborhood operators. TopoTune \citep{papillon2025topotune} introduces general topological neural networks that decompose complexes into strictly augmented Hasse graphs, letting any GNN backbone process incidence and adjacency neighborhoods as separate channels. HOPSE \citep{bernardez2026hopse} uses the same decompositions for higher-order positional encodings at linear cost. All of these works generally assume the input complex is fixed, leaving the choice of lifting as a separate, often unexamined preprocessing step.

\paragraph{The TDL Lifting Ecosystem}
A central TDL challenge is lifting point clouds and graphs into higher-order domains in a principled and meaningful way. Motivated by this, the 2024 Topological Deep Learning Challenge \citep{bernardez2024icml} catalogued more than 30 different liftings between different pairs of domains, which were later integrated into TopoBench~\citep{telyatnikov2024topobench} to standardize pipelines. While the TopoBench ecosystem broadens the lifting menu, it offers no diagnostics for the structural quality of the resulting complexes. \citet{rieck2025have} argues that such liftings may only artificially inflate model performance by increasing parameter size while relying on largely irrelevant domain structure. Only one systematic study of various liftings exists and is tied to hypergraph neural networks \citep{montagna2026lift}. Our work helps address this broader gap by introducing a general diagnostic tool capable of analyzing structural properties across any higher-order domain. On the visualization side, general-purpose network exploration tools such as Gephi \citep{bastian2009gephi} and Cytoscape \citep{shannon2003cytoscape} support interactive inspection of pairwise graphs, but no existing tool exposes the cross-rank neighborhood structure of lifted higher-order domains.



\paragraph{Graph Metrics and Model Performance}
Structural descriptors are essential for understanding when and why graph learning succeeds. Reproducibility studies revealed that arbitrary data splits and hyperparameter choices mask true model capabilities \citep{shchur2018pitfalls}, and that structure-agnostic baselines often outperform complex architectures under fair protocols \citep{errica2020fair}. \citet{dwivedi2023benchmarking} introduced datasets targeting structural motifs and over-smoothing, while synthetic frameworks like GraphWorld \citep{palowitch2022graphworld} systematically vary graph properties to study their effect on learning. On the diagnostic side, \citet{pei2020geom} identified heterophily as a primary failure mode for message passing, though \citet{platonov2023critical} showed that standard heterophily metrics poorly predict performance in practice. In discrete geometry, \citet{weber2017forman} introduced Forman-Ricci curvature as an edge-based network characteristic; subsequent work proved that negatively curved edges cause over-squashing \citep{topping2022understanding}, motivating curvature-based rewiring \citep{fesser2024mitigating} and structural encodings \citep{fesser2024effective} as corrective strategies. These results demonstrate that geometric structural metrics can provide principled, actionable signals for graph learning. TopoExplorer extends this perspective to higher-order domains by leveraging the (strictly augmented) Hasse graph representation. This allows computing and displaying classical graph metrics as well as edge-specific Forman–Ricci curvature for these higher-order domains.

\paragraph{Topological Metrics and Model Performance}
Related efforts have begun extending such diagnostics to higher-order domains. \citet{telyatnikov2023hypergraph} and \citet{sarker2024simplicialhomophily} extended the concept of homophily to hypergraphs and simplicial complexes, explicitly tying higher-order structural properties to the empirical performance of higher-order neural networks. Such metrics remain specialized and domain-specific. By representing arbitrary higher-order domains as strictly augmented Hasse graphs, TopoExplorer instead enables practitioners to compute a broad range of graph diagnostics across the full topological hierarchy.

\section{TopoExplorer: Interact and Diagnose Any Lifted Topological Dataset}
We present an interactive visual and quantitative diagnostic framework built on top of the TopoBench ecosystem \citep{telyatnikov2024topobench}. Its central objective is to expose the structural implications of a lifting $\mathcal{L}$ and its hyperparameters before any computational resources are expended on model training. By decomposing the resulting combinatorial complex $\mathcal{C}$ into its strictly augmented Hasse graphs \citep{papillon2025topotune}, the interface provides a direct view of how information will flow through each topological neighborhood.

\subsection{Interface Design}
\begin{figure}[t]
    \centering
    \vspace{-60pt}
    \includegraphics[width=\linewidth]{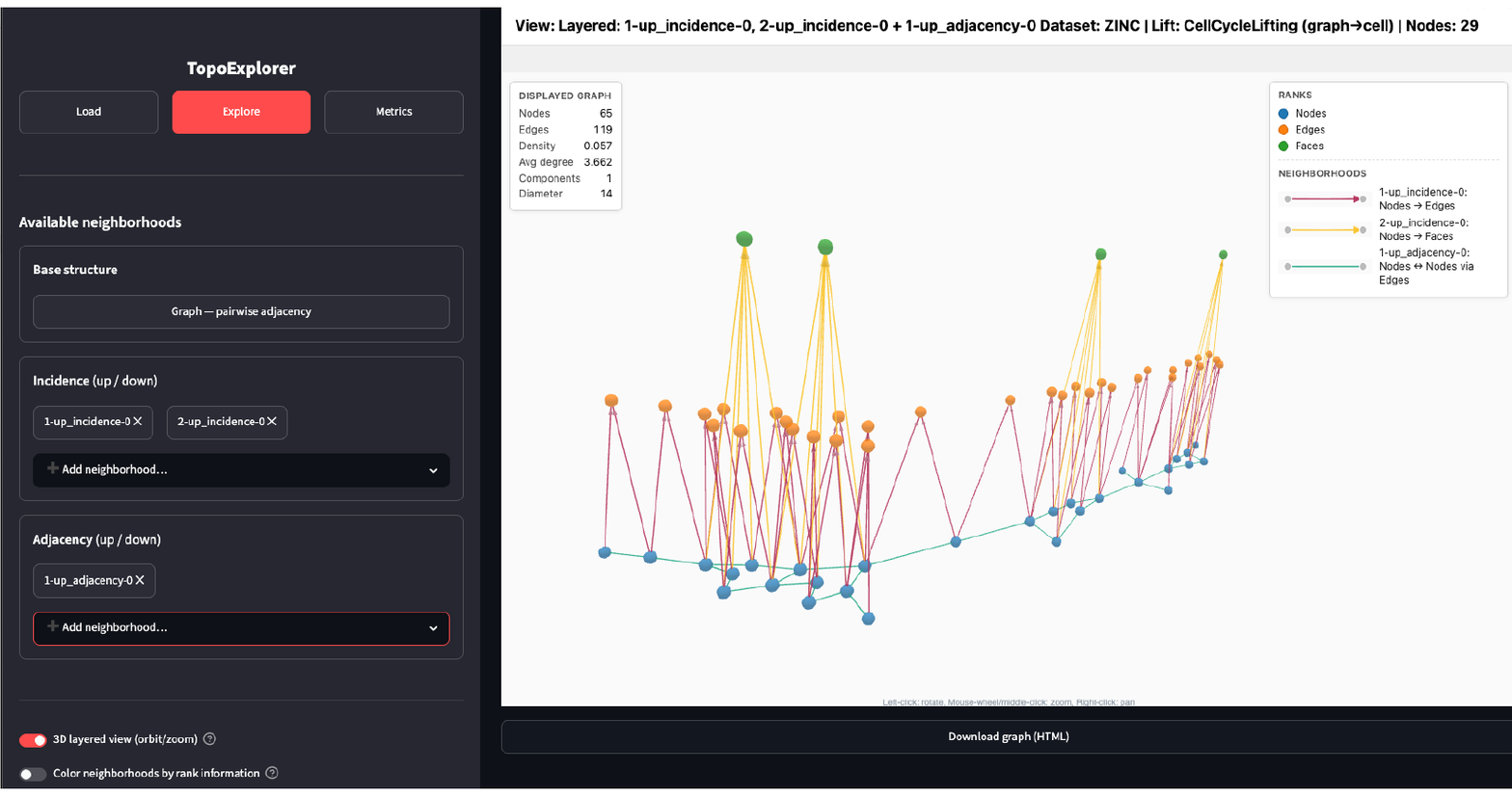}
     \vspace{-60pt}
    \caption{Neighborhood selection and representation in TopoExplorer. Example of a cellular complex from the ZINC \citep{irwin2012zinc} dataset, with blue 0-cells (nodes), orange 1-cells (edges) and green 2-cells (faces) represented in terms of Hasse graphs, where the color-coded edges encode neighborhood-specific relations. For a full-page version of this figure, we refer to Appendix \ref{app:full-page-figures}.}
    \label{fig:nbhd_example}
\end{figure}

\paragraph{Stratified rank layout.} To avoid TopoExplorer displaying confusingly dense cell intersections, every cell is represented as a distinct node, organized into hierarchical layers by rank (0-cells at the bottom, 1-cells in the middle, 2-cells at the top). Fig. \ref{fig:nbhd_example} shows an example of this construction. 

\paragraph{Neighborhood representation.} \textit{i.} Incidence neighborhoods (boundary and coboundary) are represented as directed edges \textit{between} two layers: for example, the edges connecting the bottom layer (0-cells) to the second layer (1-cells) represent $\mathcal{I}_{0 \rightarrow 1}$. If selected, TopoExplorer's incidence dashboard exclusively depicts these neighborhoods to reveal cross-rank connectivity. This shows how higher-order cells are supported by their lower-rank constituents and whether the lifting has generated orphan cells or clusters of cells that may act as sinks during message passing. \textit{ii.} Adjacency neighborhoods (both lower and upper adjacencies) are represented as edges \textit{within} one layer. If selected, TopoExplorer's adjacency dashboard, restricted to just these neighborhoods, reveals how the intra-rank, peer-to-peer connectivity evolves across and over ranks.

\vspace{55pt}
\subsection{Quantitative Signatures}
 
Each Hasse graph $G_{\mathcal{N}}$ is paired with real-time connectivity metrics to support principled architecture design, all listed and defined in Appendix \ref{app:topoexplorer-metrics}. We provide some examples here:
 
\begin{itemize}
    \item \textbf{Edge Sparsity and Density:} Near-100\% sparsity in a given neighborhood graph gives practitioners a principled basis for pruning that message-passing channel in TopoTune, saving compute without sacrificing expressivity.
    \item \textbf{Degree Statistics:} TopoExplorer summarizes the degree distribution of each Hasse graph through its average, maximum, and minimum degree, degree assortativity, and per-node degree centralities (Table~\ref{tab:topoexplorer-metrics}). Highly skewed degree statistics signal topological bottlenecks, flagging where normalization techniques or attention mechanisms may be necessary.
    \item \textbf{Component Analysis:} The number of connected components determines whether structural encodings will capture global context or remain localized.
\end{itemize}

We emphasize that these metrics carry no universal thresholds; they are intended to be interpreted comparatively, across candidate liftings, neighborhood choices, and hyperparameter configurations for a given dataset. TopoExplorer also offers an HTML export file recording all computed metrics, allowing users to record and compare values across configurations.

\subsection{Practitioner Workflow}
 
\begin{tcolorbox}[colback=gray!10, colframe=gray!50, arc=4mm, boxrule=0.5pt]
\begin{enumerate}
    \item \textbf{Select a dataset.} Choose a topological domain and dataset. The app displays descriptive metadata (task, number of features, number of classes, etc.).
    \item \textbf{Configure a lifting.} Select a target domain (hypergraph, simplicial, cell, combinatorial) and lifting method $\mathcal{L}$ from the TopoBench catalogue, together with its hyperparameters $\theta$, yielding a configuration $\mathcal{L}(\theta)$.
    \item \textbf{Load and lift.} Click \emph{Load graph} to load and cache the dataset via the TopoBench API. The transform $\mathcal{L}(\theta)$ is applied to the selected sample $G_i$, producing a complex $\mathcal{T} = (\mathcal{V}, \mathcal{C}, \mathrm{rk})$.
    \item \textbf{Select neighborhoods.} Choose one or more neighborhood types: graph adjacency, graph incidence,  higher-order adjacency (through various ranks), higher-order incidence (between various ranks). Each is represented as a Hasse graph $\mathcal{G}_{\mathcal{N}}$ that can be inspected individually or combined to compare information flow across ranks.
    \item \textbf{Adjust visualization settings.} For inductive datasets, navigate between data samples $G_i$; the lifting $\mathcal{L}(\theta)$ is reapplied automatically. Adjust display filters (minimum cell degree, maximum cells per rank) without modifying the underlying complex.
    \item \textbf{Render and export.} Explore the interactive visualization, open it in a standalone browser window, or export it as a self-contained HTML file.
\end{enumerate}
\end{tcolorbox}  

\paragraph{Beyond the hosted app.} Practitioners are not restricted to the datasets and liftings served by the hosted instance. Because TopoExplorer is released open-source, the repository can be cloned and adapted to accommodate custom datasets, liftings, and metrics, and thereby embedded within an existing experimental pipeline. Extensive documentation accompanying the code details the available metrics and their interpretation, along with the recommended diagnostic workflows.

\begin{figure}
  \centering
  \includegraphics[width=\linewidth]{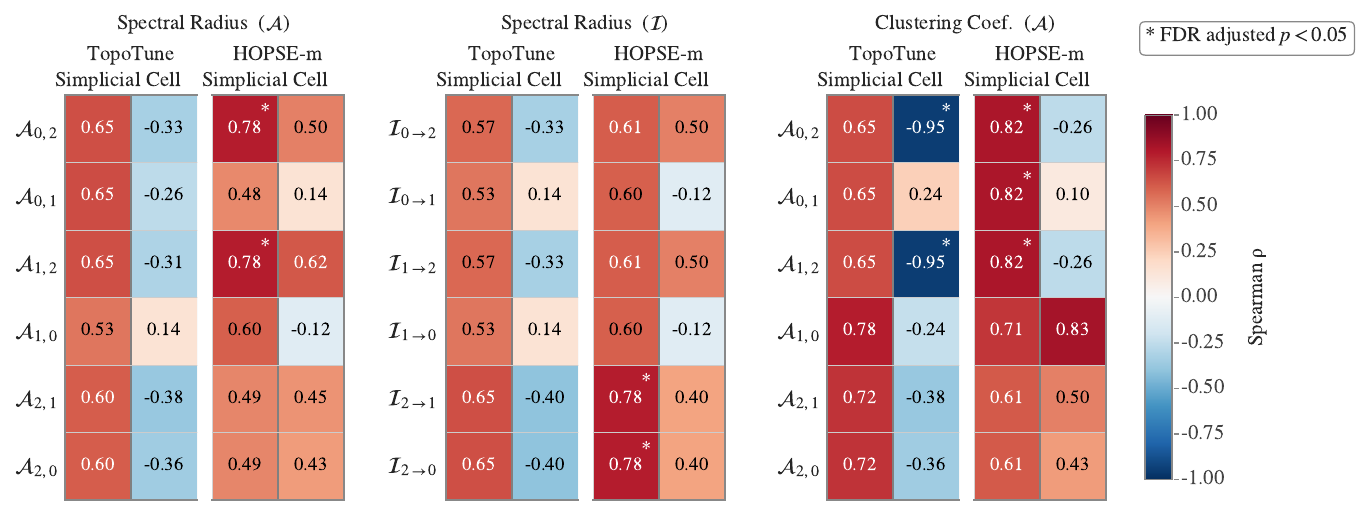}
\caption{%
  \textbf{Spearman $\rho$ between Hasse graph metrics and normalized model performance.} We explore (left) spectral radius $\lambda_{\max}$ on adjacency Hasse graphs $\mathcal{G}_{\mathcal{A}_{t,s}}$, (middle) spectral radius on incidence Hasse graphs $\mathcal{G}_{\mathcal{I}_{s\to t}}$, and (right) clustering coefficient $C$ on adjacency Hasse graphs. The simplicial and cellular domains draw from 12 and 8 datasets, respectively.
  A $*$ marks significant correlations.
}
  \label{fig:case_study_heatmap}
\end{figure}


\subsection{TopoExplorer in Action: Structural Diagnostics for Neighborhood Selection}

A central motivation for TopoExplorer is the hypothesis that the structure induced by a neighborhood operator contains useful signals about its downstream utility. Because TopoExplorer exposes the strictly augmented Hasse graph $\mathcal{G}_\mathcal{N}$ associated with any neighborhood $\mathcal{N}$, practitioners can inspect these structural properties before training a model. We evaluate this hypothesis through a case study relating (strictly augmented) Hasse graph diagnostics to model performance.

\begin{figure}[t]
  \centering
  \includegraphics[width=\linewidth]{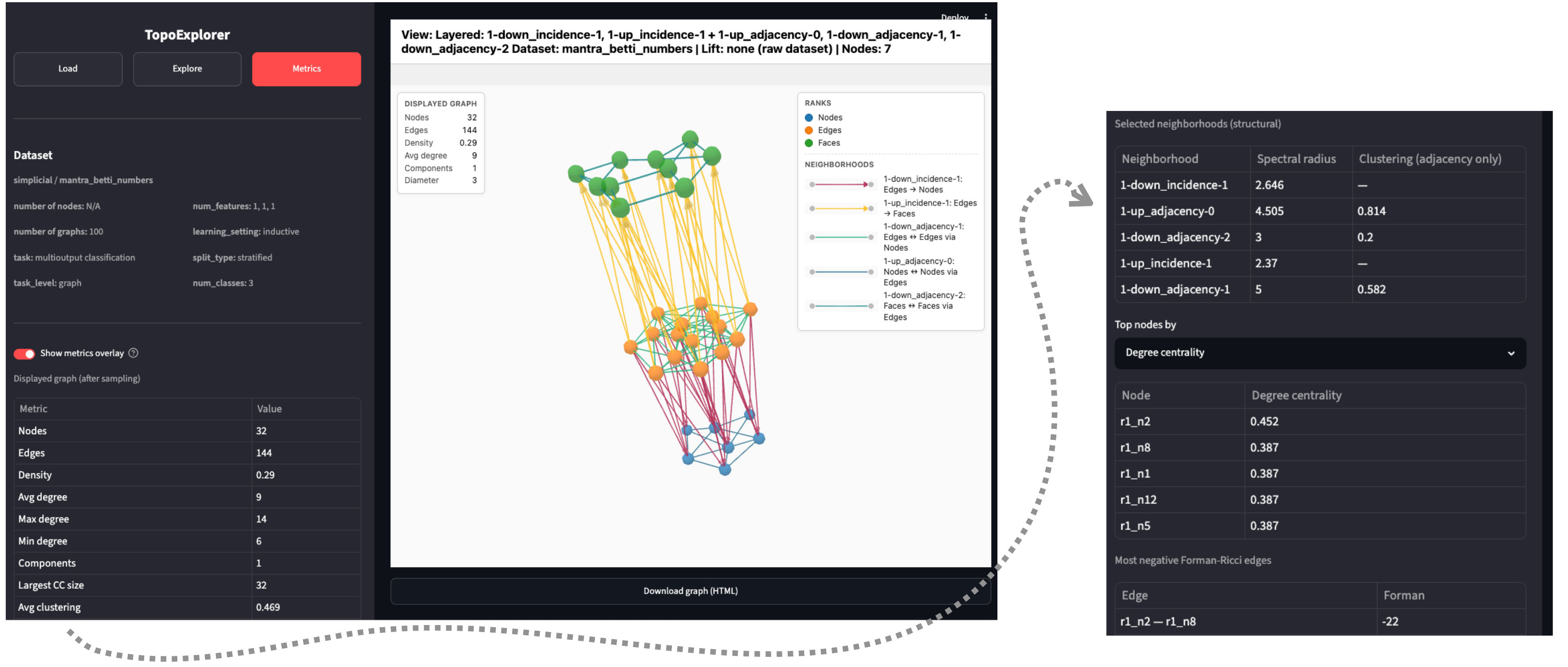}
  \caption{%
    TopoExplorer displaying an interactive rendering of MANTRA \citep{ballester2025mantra} simplicial dataset sample (see full-page version in Appendix \ref{app:full-page-figures}).
    For each of the 5 selected neighborhoods, the interface (see right panel screenshot) reports the spectral radius $\lambda_{\max}(\mathcal{G}_\mathcal{N})$ and clustering coefficient $C(\mathcal{G}_\mathcal{N})$ alongside other metrics (see full list in Appendix \ref{app:topoexplorer-metrics}).
  }
  \label{fig:topoexplorer_case_study}
\end{figure}

\paragraph{Structural metrics.}
For a neighborhood $\mathcal{N}$, the induced Hasse graph (Eq.~\ref{eq:hasse}) is amenable to standard graph analysis. We consider two descriptors available directly within TopoExplorer: (i) the \emph{spectral radius} $\lambda_{\max}(\mathcal{G}_\mathcal{N})$, which captures overall connectivity and expansion, and (ii) the \emph{clustering coefficient} $C(\mathcal{G}_\mathcal{N})$, which measures local cohesiveness. Because incidence Hasse graphs are bipartite, clustering coefficients are only defined for adjacency neighborhoods. Metrics are computed per sample and averaged at the dataset level (Figure~\ref{fig:topoexplorer_case_study}).

\paragraph{Experimental setup.}
We study whether these pre-training descriptors associate with the performance of TopoTune~\citep{papillon2025topotune} and HOPSE~\citep{bernardez2026hopse}, two architectures that process adjacency ($\mathcal{A}_{t,s}$) and incidence ($\mathcal{I}_{s \to t}$) neighborhoods as separate channels. We use the best-configuration results from the large-scale experiments on 20 benchmark datasets (12 simplicial, 8 cellular) previously reported in each of these works. We compute the Spearman correlation between each dataset-level metric and performance normalized relative to the strongest graph baseline (GCN \citep{kipf2017semi}, GIN \citep{xu2018how}, GAT \citep{velickovic2017graph}). Correlations are computed separately for simplicial and cellular domains, with Benjamini-Hochberg correction ($\alpha=0.05$).

\paragraph{Observations.}
Figure~\ref{fig:case_study_heatmap} reveals two key findings. First, the strongest associations are concentrated in neighborhoods involving higher-order cells, particularly 2-cell adjacencies and incidence neighborhoods. This provides correlational evidence that higher-order connectivity patterns are associated with downstream performance, though establishing that these signals reliably improve design choices requires further controlled experiments. Second, simple Hasse-graph descriptors correlate with model gains. On simplicial datasets, larger spectral radius and clustering coefficient generally associate with stronger improvements, especially for HOPSE-m. Positive trends are also observed for incidence neighborhoods. While relationships are weaker and less consistent on cellular datasets, the overall results suggest that neighborhood structure contains predictive signals about neighborhood utility before training.

\paragraph{Pre-training utility.} These descriptors are directly available within TopoExplorer. As shown in Figure~\ref{fig:topoexplorer_case_study}, selecting a neighborhood automatically renders its Hasse graph together with its structural metrics. Beyond visualization, TopoExplorer therefore provides interpretable diagnostics that can help guide neighborhood selection prior to model training.

\section{Conclusion and Future Work}

We present TopoExplorer, an interactive diagnostic framework that makes the structural properties of strictly augmented Hasse graphs $\mathcal{G}_\mathcal{N}$ visible and measurable before any model training.
By exposing neighborhood-level metrics such as spectral radius and clustering coefficient directly in the application interface, TopoExplorer bridges the gap between the lifting step and the neighborhood operator selection that conditions the performance of general TDL architectures.
Our case study demonstrates that these pre-training structural descriptors carry domain- and model-dependent associations with downstream performance, providing interpretable signals to inform the \emph{lift-look-design-train} workflow we advocate. Future work includes broadening the space of structural descriptors surfaced by TopoExplorer, including geometric and feature-based metrics across all neighborhood types, and examining their predictive value for a wider range of liftings and architectures.

\newpage

\section*{Acknowledgements}
Mathilde Papillon acknowledges funding from Noyce Foundation and NSF Career 240158. 
Guillermo Bernárdez acknowledges funding from NSF 2602079 and Arlequin AI. 
Marco Montagna ackowledges support from Sapienza grant RG123188B3EF6A80 (CENTS).
R\'emi Devaux and Antoine Jardin gratefully acknowledge the NOESIS project, part of the "Pionniers de l'IA" program, for providing the resources that supported this work on TopoExplorer.
Nina Miolane acknowledges funding from the NSF Career 2240158.

\bibliography{references}

@misc{bernardez2026hopse,
      title={HOPSE: Scalable Higher-Order Positional and Structural Encoder for Combinatorial Representations}, 
      author={Guillermo Bernárdez and Marco Montagna and Louis Van Langendonck and Martin Carrasco and Amirreza Akbari and Louisa Cornelis and Mathilde Papillon and Pere Barlet-Ros and Nina Miolane and Lev Telyatnikov},
      year={2026},
      eprint={2505.15405},
      archivePrefix={arXiv},
      primaryClass={cs.LG},
      url={https://arxiv.org/abs/2505.15405}, 
}

@article{shannon2003cytoscape,
  title={Cytoscape: a software environment for integrated models of biomolecular interaction networks},
  author={Shannon, Paul and Markiel, Andrew and Ozier, Owen and Baliga, Nitin S and Wang, Jonathan T and Ramage, Daniel and Amin, Nada and Schwikowski, Benno and Ideker, Trey},
  journal={Genome research},
  volume={13},
  number={11},
  pages={2498--2504},
  year={2003},
  publisher={Cold Spring Harbor Lab}
}

@article{dwivedi2023benchmarking,
  title={Benchmarking graph neural networks},
  author={Dwivedi, Vijay Prakash and Joshi, Chaitanya K and Luu, Anh Tuan and Laurent, Thomas and Bengio, Yoshua and Bresson, Xavier},
  journal={Journal of Machine Learning Research},
  volume={24},
  number={43},
  pages={1--48},
  year={2023}
}

@inproceedings{pei2020geom,
title={Geom-GCN: Geometric Graph Convolutional Networks},
author={Hongbin Pei and Bingzhe Wei and Kevin Chen-Chuan Chang and Yu Lei and Bo Yang},
booktitle={International Conference on Learning Representations},
year={2020},
url={https://openreview.net/forum?id=S1e2agrFvS}
}

@article{shchur2018pitfalls,
  title={Pitfalls of Graph Neural Network Evaluation},
  author={Shchur, Oleksandr and Mumme, Maximilian and Bojchevski, Aleksandar and G{\"u}nnemann, Stephan},
  journal={Relational Representation Learning Workshop, NeurIPS 2018},
  year={2018}
}

@inproceedings{errica2020fair,
title={A Fair Comparison of Graph Neural Networks for Graph Classification},
author={Federico Errica and Marco Podda and Davide Bacciu and Alessio Micheli},
booktitle={International Conference on Learning Representations},
year={2020},
url={https://openreview.net/forum?id=HygDF6NFPB}
}

@article{bernardez2024icml,
  title={ICML Topological Deep Learning Challenge 2024: Beyond the Graph Domain},
  author={Bern{\'a}rdez, Guillermo and Telyatnikov, Lev and Montagna, Marco and Baccini, Federica and Papillon, Mathilde and Ferriol-Galm{\'e}s, Miquel and Hajij, Mustafa and Papamarkou, Theodore and Bucarelli, Maria Sofia and Zaghen, Olga and others},
  journal={Proceedings of Machine Learning Research},
  volume={251},
  pages={420--428},
  year={2024},
  publisher={ML Research Press}
}

@inproceedings{bastian2009gephi,
  title={Gephi: an open source software for exploring and manipulating networks},
  author={Bastian, Mathieu and Heymann, Sebastien and Jacomy, Mathieu},
  booktitle={Proceedings of the international AAAI conference on web and social media},
  volume={3},
  number={1},
  pages={361--362},
  year={2009}
}

@inproceedings{papillon2025topotune,
  title={TopoTune: A Framework for Generalized Combinatorial Complex Neural Networks},
  author={Papillon, Mathilde and Bernardez, Guillermo and Battiloro, Claudio and Miolane, Nina},
  booktitle={International Conference on Machine Learning},
  pages={47924--47952},
  year={2025},
  organization={PMLR}
}

@article{papillon2023architectures,
  title={Architectures of topological deep learning: A survey of message-passing topological neural networks},
  author={Papillon, Mathilde and Sanborn, Sophia and Hajij, Mustafa and Miolane, Nina},
  journal={arXiv preprint arXiv:2304.10031},
  year={2023}
}

@article{
telyatnikov2024topobench,
title={TopoBench: A Framework for Benchmarking Topological Deep Learning},
author={Lev Telyatnikov and Guillermo Bernardez and Marco Montagna and Mustafa Hajij and Martin Carrasco and Pavlo Vasylenko and Mathilde Papillon and Ghada Zamzmi and Michael T Schaub and Jonas Verhellen and Pavel Snopov and Bertran Miquel-Oliver and Manel Gil-Sorribes and Alexis Molina and VICTOR GUALLAR and Theodore Long and Julian Suk and Patryk Rygiel and Alexander V Nikitin and Giordan Escalona and Michael Banf and Dominik Filipiak and Liliya Imasheva and Max Schattauer and Alvaro L. Martinez and Halley Fritze and Marissa Masden and Valentina S{\'a}nchez and Manuel Lecha and Andrea Cavallo and Claudio Battiloro and Matthew Piekenbrock and Mauricio Tec and George Dasoulas and Nina Miolane and Simone Scardapane and Theodore Papamarkou},
journal={Journal of Data-centric Machine Learning Research},
year={2025},
url={https://openreview.net/forum?id=07sTzyEVtY},
note={}
}

@phdthesis{bodnar2023topological,
  title={Topological deep learning: graphs, complexes, sheaves},
  author={Bodnar, Cristian},
  school={University of Cambridge}, 
  year={2023}
}

@inproceedings{
platonov2023critical,
title={A critical look at the evaluation of {GNN}s under heterophily: Are we really making progress?},
author={Oleg Platonov and Denis Kuznedelev and Michael Diskin and Artem Babenko and Liudmila Prokhorenkova},
booktitle={The Eleventh International Conference on Learning Representations },
year={2023},
url={https://openreview.net/forum?id=tJbbQfw-5wv}
}

@article{bodnar2021weisfeiler,
	title        = {Weisfeiler and {Lehman} {Go} {Cellular}: {CW} {Networks}},
	author       = {Bodnar, Cristian and Frasca, Fabrizio and Otter, Nina and Wang, Yuguang and Lio, Pietro and Montufar, Guido F and Bronstein, Michael},
	year         = 2021,
	journal      = {Advances in Neural Information Processing Systems},
	volume       = 34,
	pages        = {2625--2640}
}

@inproceedings{velickovic2017graph,
title={Graph Attention Networks},
author={Petar Veličković and Guillem Cucurull and Arantxa Casanova and Adriana Romero and Pietro Liò and Yoshua Bengio},
booktitle={International Conference on Learning Representations},
year={2018},
url={https://openreview.net/forum?id=rJXMpikCZ},
}

@inproceedings{xu2018how,
	title        = {How Powerful are Graph Neural Networks?},
	author       = {Keyulu Xu and Weihua Hu and Jure Leskovec and Stefanie Jegelka},
	year         = 2019,
	booktitle    = {International Conference on Learning Representations},
	url          = {https://openreview.net/forum?id=ryGs6iA5Km}
}

@inproceedings{topping2022understanding,
	title        = {Understanding over-squashing and bottlenecks on graphs via curvature},
	author       = {Jake Topping and Francesco Di Giovanni and Benjamin Paul Chamberlain and Xiaowen Dong and Michael M. Bronstein},
	year         = 2022,
	booktitle    = {International Conference on Learning Representations},
	url          = {https://openreview.net/forum?id=7UmjRGzp-A}
}

@inproceedings{ebli2020simplicial,
	title        = {Simplicial Neural Networks},
	author       = {S. Ebli and M. Defferrard and G. Spreemann},
	year         = 2020,
	booktitle    = {Advances in Neural Information Processing Systems Workshop on Topological Data Analysis and Beyond}
}

@inproceedings{kipf2017semi,
	title        = {Semi-Supervised Classification with Graph Convolutional Networks},
	author       = {Kipf, Thomas N. and Welling, Max},
	year         = 2017,
	booktitle    = {International Conference on Learning Representations (ICLR)}
}

@inproceedings{weber2018coarse,
  title     = {Detecting the Coarse Geometry of Networks},
  author    = {Weber, Melanie and Jost, J{\"u}rgen and Saucan, Emil},
  booktitle = {NeurIPS Workshop on Relational Representation Learning},
  year      = {2018},
}

@article{watts1998collective,
  title={Collective dynamics of ‘small-world’networks},
  author={Watts, Duncan J and Strogatz, Steven H},
  journal={nature},
  volume={393},
  number={6684},
  pages={440--442},
  year={1998},
  publisher={Nature Publishing Group}
}

@inproceedings{fesser2024effective,
	author = {Fesser, Lukas and Weber, Melanie},
	booktitle = {International Conference on Learning Representations},
	editor = {B. Kim and Y. Yue and S. Chaudhuri and K. Fragkiadaki and M. Khan and Y. Sun},
	pages = {22571--22589},
	title = {Effective Structural Encodings via Local Curvature Profiles},
	url = {https://proceedings.iclr.cc/paper_files/paper/2024/file/6188c02ccc16a7587716de2efd754033-Paper-Conference.pdf},
	volume = {2024},
	year = {2024}}

@inproceedings{fesser2024mitigating,
  title={Mitigating over-smoothing and over-squashing using augmentations of forman-ricci curvature},
  author={Fesser, Lukas and Weber, Melanie},
  booktitle={Learning on Graphs Conference},
  pages={19--1},
  year={2024},
  organization={PMLR}
}

@article{rieck2025have,
  title={Have Graph -- Will Lift? The Case for Higher-Order Benchmarks},
  author={Rieck, Bastian},
  journal={Proceedings of Workshop on Geometry, Topology, and Machine Learning (GTML)},
  year={2025}
}

@inproceedings{palowitch2022graphworld,
  title={Graphworld: Fake graphs bring real insights for gnns},
  author={Palowitch, John and Tsitsulin, Anton and Mayer, Brandon and Perozzi, Bryan},
  booktitle={Proceedings of the 28th ACM SIGKDD conference on knowledge discovery and data mining},
  pages={3691--3701},
  year={2022}
}

@article{irwin2012zinc,
	title        = {{ZINC}: a free tool to discover chemistry for biology},
	author       = {Irwin, John J and Sterling, Teague and Mysinger, Michael M and Bolstad, Erin S and Coleman, Ryan G},
	year         = 2012,
	journal      = {Journal of Chemical Information and Modeling},
	volume       = 52,
	number       = 7,
	pages        = {1757--1768}
}

@article{telyatnikov2023hypergraph,
  title={Hypergraph neural networks through the lens of message passing: a common perspective to homophily and architecture design},
  author={Telyatnikov, Lev and Bucarelli, Maria Sofia and Bernardez, Guillermo and Zaghen, Olga and Scardapane, Simone and Lio, Pietro},
  journal={arXiv preprint arXiv:2310.07684},
  year={2023}
}

@book{hajij2024topologicalbook,
  title = {Topological Deep Learning: Going Beyond Graph Data},
  author = {Mustafa Hajij and Theodore Papamarkou and Ghada Zamzmi and Karthikeyan Natesan Ramamurthy and Tolga Birdal and Michael T. Schaub},
  year = {2024},
  url = {http://tdlbook.org},
  publisher = {Online},
  note = {Published online on August 6, 2024}
}

@inproceedings{
ballester2025mantra,
title={{MANTRA}: The Manifold Triangulations Assemblage},
author={Rub{\'e}n Ballester and Ernst R{\"o}ell and Daniel Bin Schmid and Mathieu Alain and Sergio Escalera and Carles Casacuberta and Bastian Rieck},
booktitle={The Thirteenth International Conference on Learning Representations},
year={2025},
url={https://openreview.net/forum?id=X6y5CC44HM}
}

@misc{hajij2023topological,
	title        = {Topological Deep Learning: Going Beyond Graph Data},
	author       = {Mustafa Hajij and Ghada Zamzmi and Theodore Papamarkou and Nina Miolane and Aldo Guzmán-Sáenz and Karthikeyan Natesan Ramamurthy and Tolga Birdal and Tamal K. Dey and Soham Mukherjee and Shreyas N. Samaga and Neal Livesay and Robin Walters and Paul Rosen and Michael T. Schaub},
	year         = 2023,
	eprint       = {2206.00606},
	archiveprefix = {arXiv},
	primaryclass = {cs.LG}
}

@inproceedings{montagna2026lift,
  title={Lift Me Up: The Impact of Liftings on Hypergraph Neural Networks},
  author={Montagna, Marco and Scardapane, Simone and Telyatnikov, Lev},
  booktitle={ICLR 2026 GRAM Workshop},
  year={2026}
}

@article{weber2017forman,
  title={Characterizing complex networks with Forman-Ricci curvature and associated geometric flows},
  author={Weber, Melanie and Saucan, Emil and Jost, J{\"u}rgen},
  journal={Journal of Complex Networks},
  volume={5},
  number={4},
  pages={527--550},
  year={2017},
  publisher={Oxford University Press}
}

@article{sarker2024simplicialhomophily, title={Higher-order homophily on simplicial complexes}, volume={121}, DOI={https://doi.org/10.1073/pnas.2315931121}, number={12}, journal={Proceedings of the National Academy of Sciences of the United States of America}, publisher={National Academy of Sciences}, author={Sarker, Arnab and Northrup, Natalie and Jadbabaie, Ali}, year={2024}, month={Mar} }
\newpage
\appendix

\section{Definition of a Neighborhood} \label{app:nbhds}
Here we provide the complete mathematical definition for neighborhoods of a combinatorial complex $\mathcal{T}$. The incidence neighborhood of a $t$-cell $\sigma\in\mathcal{T}$ with respect to a rank $s$ is defined as 
    \begin{equation}\label{eq:incidences}
    \begin{split}
        \mathcal{I}_{s\to t}(\sigma) =
        \left\{
        \begin{array}{ll}
             \{\tau \in \mathcal{C} \, | \, \textrm{rk}(\tau) = s, \sigma \subset \tau \} & \text{ if $t < s$;} \\
             \{\tau \in \mathcal{C} \, | \, \textrm{rk}(\tau) = s, \tau \subset \sigma\} & \text{ if $t > s$.}
        \end{array}
        \right.
    \end{split}
    \end{equation}
A $s$-cell $\tau\in\mathcal{T}$ is said to be co-incident of $\sigma$ if it contains $\sigma$ and $s>t$; analogously, a $s$-cell $\tau\in\mathcal{T}$ is incident of $\sigma$ if it is contained by $\sigma$ and $s<t$.
Incidences in turn induce adjacency neighborhoods of cells of the same rank as follows:

\begin{equation}\label{eq:adjacencies}
\begin{split}
    \mathcal{A}_{s,t}(\sigma) = \{\tau \in \mathcal{C} \, | \, \textrm{rk}(\tau) = t, \mathcal{I}_{s\to t}(\sigma) \cap \mathcal{I}_{s\to t}(\tau) \neq \emptyset \}
\end{split}
\end{equation}
In this case, two $t$-cells $\sigma$ and $\tau$ are said to be adjacent if both are contained in a $s$-cell $\delta\in\mathcal{T}$.

\section{TopoExplorer metrics}
\label{app:topoexplorer-metrics}

TopoExplorer reports structural quantities on the graph loaded and configured in its visualization panel.
Unless noted otherwise, complex- and node-level statistics are computed on the \emph{displayed} graph, i.e. the subgraph rendered after dataset loading, optional lifting, and user-controlled sampling. They are evaluated on an undirected view so that the same definitions apply to plain adjacency, incidence/bipartite, and layered neighborhood displays.
Neighborhood-level statistics, by contrast, are computed on the full neighborhood matrix associated with each selected view (e.g., rank-$k$ adjacency or cross-rank incidence), before sampling is applied.
Table~\ref{tab:topoexplorer-metrics} summarizes the metrics exposed in the \emph{Metrics} tab; entries marked as advanced are optional and may be approximated on large graphs (e.g., betweenness via $k$-node sampling above ${\sim}800$ nodes).
Edge-local quantities such as Forman-Ricci curvature and edge betweenness appear in edge tooltips and in summary tables for extreme edges.

\section{Full-Page Screenshots of TopoExplorer}
\label{app:full-page-figures}

For legibility, we reproduce the two interface screenshots from the main text at
full-page scale.
\clearpage

\begin{table}[H]
\centering
\caption{Metrics provided by TopoExplorer's \emph{Metrics} tab.
Complex- and node-level quantities are computed on the \emph{displayed} graph (after sampling and lifting), treated as undirected.
Neighborhood-level quantities are computed on the full neighborhood matrix for each selected view, before sampling.}
\label{tab:topoexplorer-metrics}
\small
\begin{tabular}{@{}p{0.28\linewidth}p{0.66\linewidth}@{}}
\toprule
\textbf{Metric} & \textbf{Definition} \\
\midrule
\multicolumn{2}{@{}l}{\textit{Complex-level metrics (displayed graph)}} \\
\addlinespace
Nodes & Number of vertices in the displayed graph. \\
Edges & Number of edges in the displayed graph. \\
Density & Edge density $m / \binom{n}{2}$ for $n>1$; $0$ if $n \le 1$. \\
Avg degree & Mean vertex degree over all nodes. \\
Max degree & Maximum vertex degree. \\
Min degree & Minimum vertex degree. \\
Components & Number of connected components (undirected view). \\
Largest CC size & Number of nodes in the largest connected component. \\
Avg clustering & Global average local clustering coefficient \citep{watts1998collective}. \\
Transitivity & Global clustering coefficient (fraction of closed triples among connected triples). \\
Diameter & Maximum eccentricity in the largest connected component; omitted if the graph exceeds size limits. \\
Degree assortativity & Pearson correlation of degrees at edge endpoints; available when advanced metrics are enabled. \\
\addlinespace
\midrule
\multicolumn{2}{@{}l}{\textit{Neighborhood-level metrics (selected neighborhoods)}} \\
\addlinespace
Spectral radius & For each selected neighborhood matrix: largest $|\lambda|$ if square (symmetric adjacency), otherwise largest singular value. Computed on the full matrix, not the sampled display. \\
Clustering (adjacency only) & Average local clustering coefficient of the undirected graph induced by a square adjacency neighborhood. \\
\addlinespace
\midrule
\multicolumn{2}{@{}l}{\textit{Node-level metrics (displayed graph)}} \\
\addlinespace
Degree & Number of neighbors of the node (undirected view). \\
Degree centrality & Normalized degree $k/(n-1)$. \\
Clustering & Local clustering coefficient of the node. \\
Betweenness & Betweenness centrality; for graphs with ${>}800$ nodes, approximated by $k$-node sampling (advanced). \\
Closeness & Closeness centrality based on shortest-path distances (advanced). \\
Eccentricity & Maximum shortest-path distance from the node to any other node in its connected component (advanced). \\
Eigenvector centrality & Centrality proportional to the sum of neighbors' centralities (advanced). \\
PageRank & Stationary score under a random-walk model with damping (advanced). \\
Forman-Ricci & Per-edge curvature on the displayed graph: $4 - \deg(u) - \deg(v)$ under unit edge weights \citep{weber2018coarse}. \\
Edge betweenness & Betweenness centrality of an edge (advanced). \\
\bottomrule
\end{tabular}
\end{table}

\begin{figure}
  \centering
  \includegraphics[angle=90,height=0.92\textheight]{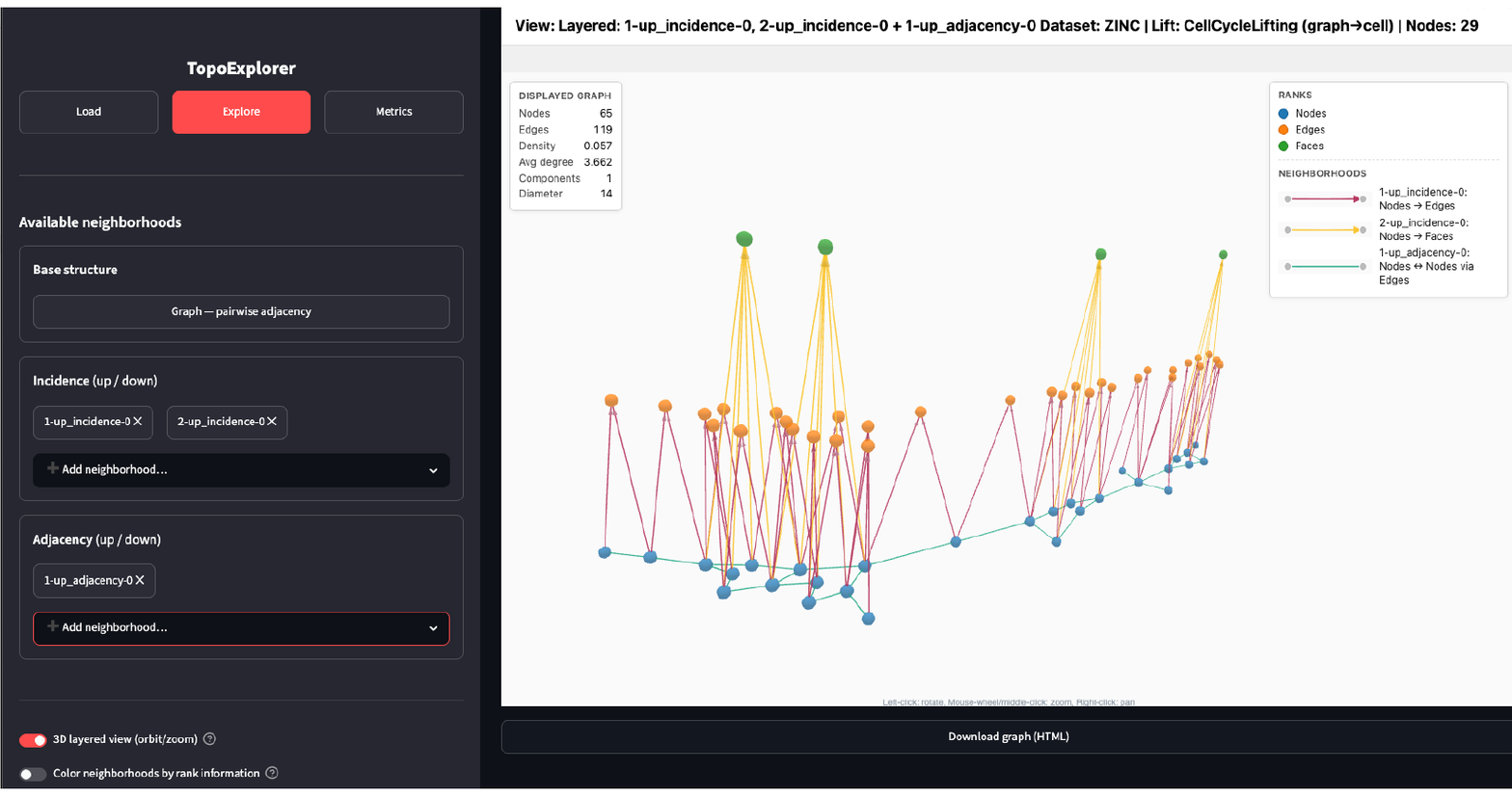}
  \caption{Full-page version of Figure~\ref{fig:nbhd_example}: neighborhood
    selection and representation in TopoExplorer on the ZINC
    \citep{irwin2012zinc} dataset}
  \label{fig:nbhd_example_large}
\end{figure}

\begin{figure}
  \centering
  \includegraphics[angle=90,height=0.92\textheight]{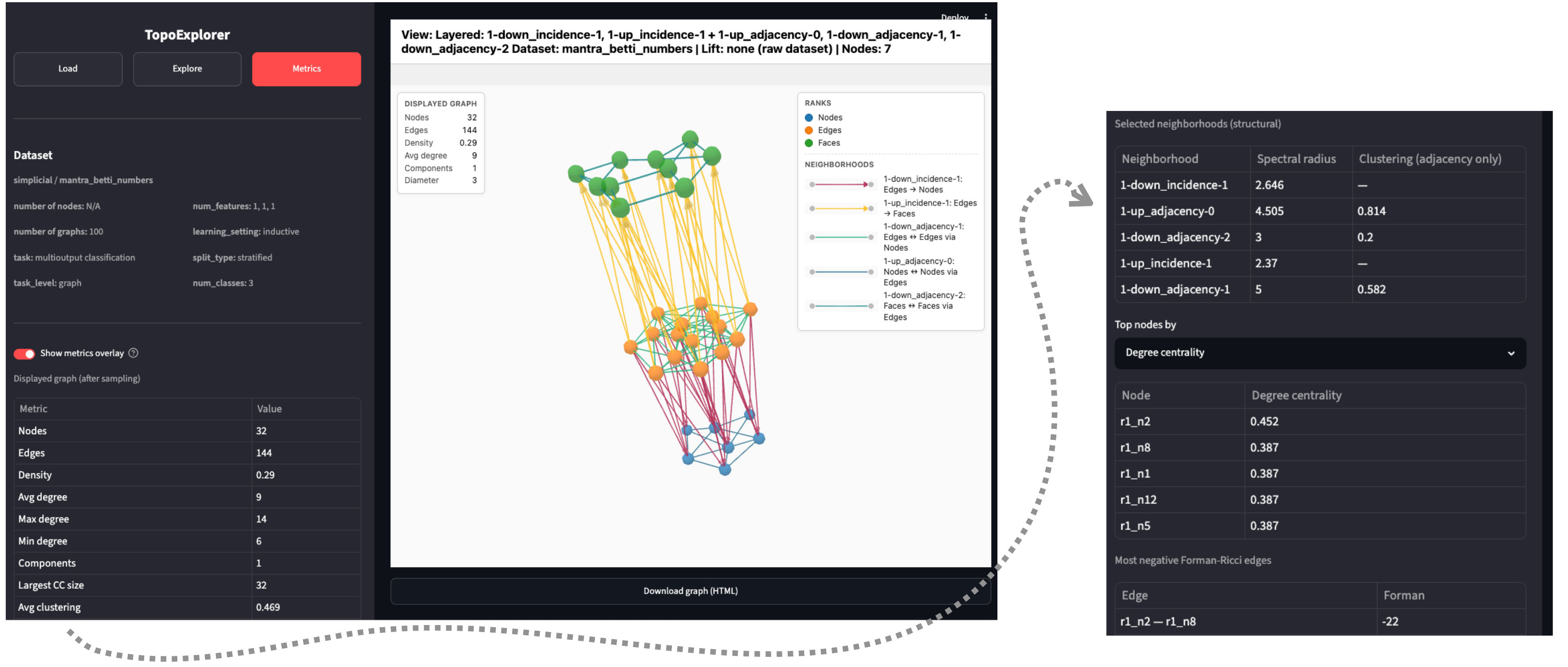}
  \caption{Full-page version of Figure~\ref{fig:topoexplorer_case_study}:
    TopoExplorer on a MANTRA \citep{ballester2025mantra} simplicial dataset
    sample, with the metrics panel enlarged.}
  \label{fig:topoexplorer_case_study_large}
\end{figure}
\end{document}